\def\year{2020}\relax
\documentclass[letterpaper]{article} 
\usepackage{aaai20}  
\usepackage{algorithm}
\usepackage{algorithmic}
\usepackage{multirow}
\usepackage{times}  
\usepackage{helvet} 
\usepackage{courier}  
\usepackage[hyphens]{url}  
\usepackage{graphicx} 
\usepackage{graphicx}  
\usepackage{subfigure}
\title{Weakly-Supervised Video Moment Retrieval\\ via Semantic Completion Network}
\author{Zhijie Lin\textsuperscript{\rm 1}, Zhou Zhao\textsuperscript{\rm 1}\thanks{Corresponding author.}, 
Zhu Zhang\textsuperscript{\rm 1},
{\bf \Large Qi Wang\textsuperscript{\rm 2}, Huasheng Liu\textsuperscript{\rm 2},} \\ 
\textsuperscript{\rm 1}College of Computer Science, Zhejiang University, Hangzhou, China, \\
\textsuperscript{\rm 2}Alibaba Inc., China\\
\{linzhijie, zhaozhou, zhangzhu\}@zju.edu.cn,
\{wq140362,fangkong.lhs\}@alibaba-inc.com}
\begin{document}


\maketitle

\begin{abstract}
Video moment retrieval is to search the moment that is most relevant to the given natural language
query.
Existing methods are mostly trained in a fully-supervised setting, which requires 
the full annotations of temporal boundary for each query.
However, manually labeling the annotations is actually time-consuming and expensive.
In this paper, we propose a novel weakly-supervised moment retrieval framework
requiring only coarse video-level annotations for training.
Specifically, we devise a proposal generation module that aggregates the context information to
generate and score all candidate proposals in one single pass.
We then devise an algorithm that considers both exploitation and exploration to select top-K proposals.
Next, we build a semantic completion module to measure the semantic similarity between the selected proposals and query, 
compute reward and provide feedbacks to the proposal generation module for scoring refinement.
Experiments on the ActivityCaptions 
and Charades-STA demonstrate the effectiveness of our proposed method.
\end{abstract}

\section{Introduction}
\begin{figure}[t]
\centering
\includegraphics[width=1.0\columnwidth]{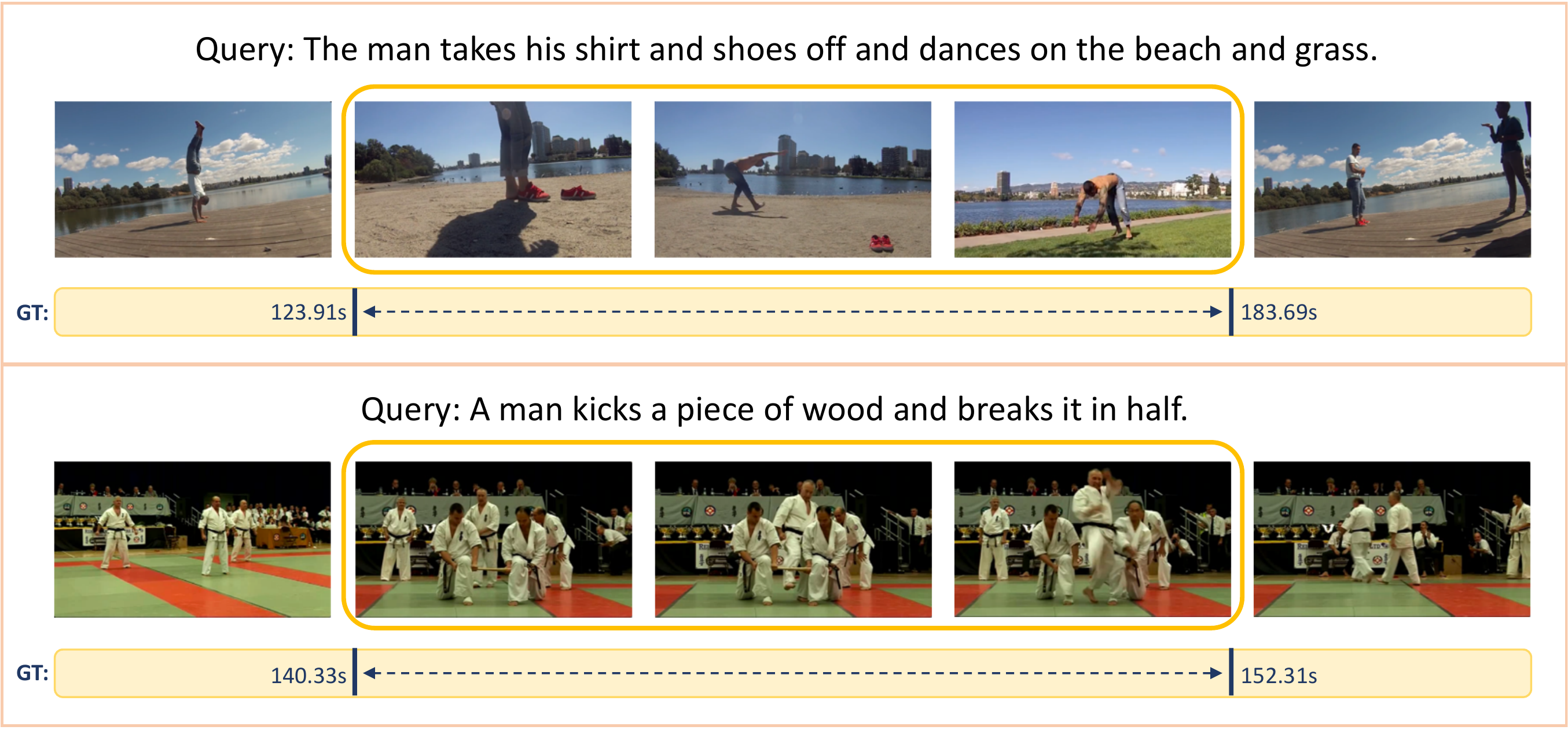} 
\caption{Examples of video moment retrieval: search the temporal boundary 
of the most relevant moment in video according to the given natural language query.}
\label{fig:example1}
\end{figure}

Video moment retrieval, a key topic in information retrieval and computer vision, has attracted more and more interests
in recent years~\cite{gao2017tall,hendricks2017localizing}. 
As two examples in Figure~\ref{fig:example1} show,
according to a given natural language query, moment retrieval aims to locate the temporal boundary of 
the most related moment in the video, which can help us quickly filter out useless contents in the video.
More accurate moment retrieval requires sufficient understanding of both the video and the query, 
which makes it a challenging task. 
Although recent works~\cite{chen2018temporally,zhang2019cross,zhang2019man} has achieved good results, 
they are mostly trained in a fully-supervised setting, which requires the full annotations of temporal boundary
for each video.
However, manually labeling the ground truth temporal boundaries is time-consuming and expensive, requiring a large amount
of human labor. 
Moreover, considering an untrimmed video contains multiple consecutive temporal activities, it can be difficult to mark
the boundaries accurately, which produces ambiguity and noise in training data.
Relatively, it is much easier to obtain coarse descriptions of a video without marking the temporal boundaries, such as the captions
of videos in YouTube. 
This motivates us to develop a weakly-supervised method for moment retrieval that needs only coarse video-level 
annotations for training.
Existing weakly-supervised method in ~\cite{mithun2019weakly} proposes to learn a joint visual-text
embedding, and utilizes the latent alignment produced by 
intermediate Text-Guided Attention~(TGA) to localize the target moment.
However, the latent attention weights without extra supervision usually 
focus on the most discriminative but small regions~\cite{singh2017hide}
instead of covering complete regions.
To deal with these issues, in this paper, 
we devise a novel weakly-supervised Semantic Completion Network~(SCN) 
including proposal generation and selection, semantic completion for semantic similarity estimation
and scoring refinement.

Firstly, rather than localizing the most relevant moment relying on the ambiguous attention weights,
we extract the semantically important proposals through a proposal generation module.
Further than method in ~\cite{gao2017tall} that treats each candidate proposal separately, 
we leverage the cross-modal fusion representations of video and
query to score all the candidate proposals sampled at different scales
in a single pass, which makes full use of context information for scoring
other proposals.

With a large set of densely sampled proposals, we then devise an algorithm 
that considers both exploitation and exploration to select top-K proposals.
Concretely, we first rank all candidate proposals
based on their corresponding confidence scores. 
Further than just selecting the proposals with high confidence score based on Non Maximum Suppression~(NMS), 
to encourage full exploration,
we select next proposal randomly with a decay possibility,
which is helpful for finding potentially good proposals
and giving more accurate confidence scores for proposals.
At the beginning of training, we tend to select next proposal randomly for exploration.
As the model converges gradually, proposals with high confidence score are chosen more often for exploitation.

To explicitly model the scoring of proposal generation module
rather than rely on attention weights without extra supervision~\cite{mithun2019weakly},
we are supposed to further measure the semantic similarity 
between the selected proposals and query for scoring refinement.
Inspired by the success of recent works about masked language model~\cite{devlin2019bert,song2019mass,wang190409408},
we design a novel semantic completion module that predicts the important words~(e.g. noun, verb) that are masked
according to the given visual context.
In detail, by masking the important words in the decoder side, SCN forces the decoder rely on the visual context
to reconstruct the query and the most semantically matching proposal can provide enough information to
predict the key words.
Then with the evaluation results given by semantic completion module, 
we compute reward for each proposal based on the reconstruction loss 
and formulate a rank loss to encourage the proposal generation module 
to give higher confidence score for those proposals with greater rewards.

In total, the main contributions of our work are listed as follows:
\begin{itemize}
    \item We propose a novel weakly-supervised moment retrieval framework 
    requiring only coarse annotations for training, 
    and experiments on two datasets: ActivityCaptions~\cite{caba2015activitynet} 
    and Charades-STA~\cite{gao2017tall} demonstrate the effectiveness of our method.
    \item We build a proposal generation module to score all candidate proposals in a single pass
    and formulate a rank loss for scoring refinement.
    \item We devise an algorithm for top-K proposals selection that encourages both exploitation
    and exploration.
    \item We design a novel semantic completion module that predicts the important words that are masked
    according to the given visual context for semantic similarity estimation.
\end{itemize}

\begin{figure*}[t]
    \centering
    \includegraphics[width=2.0\columnwidth]{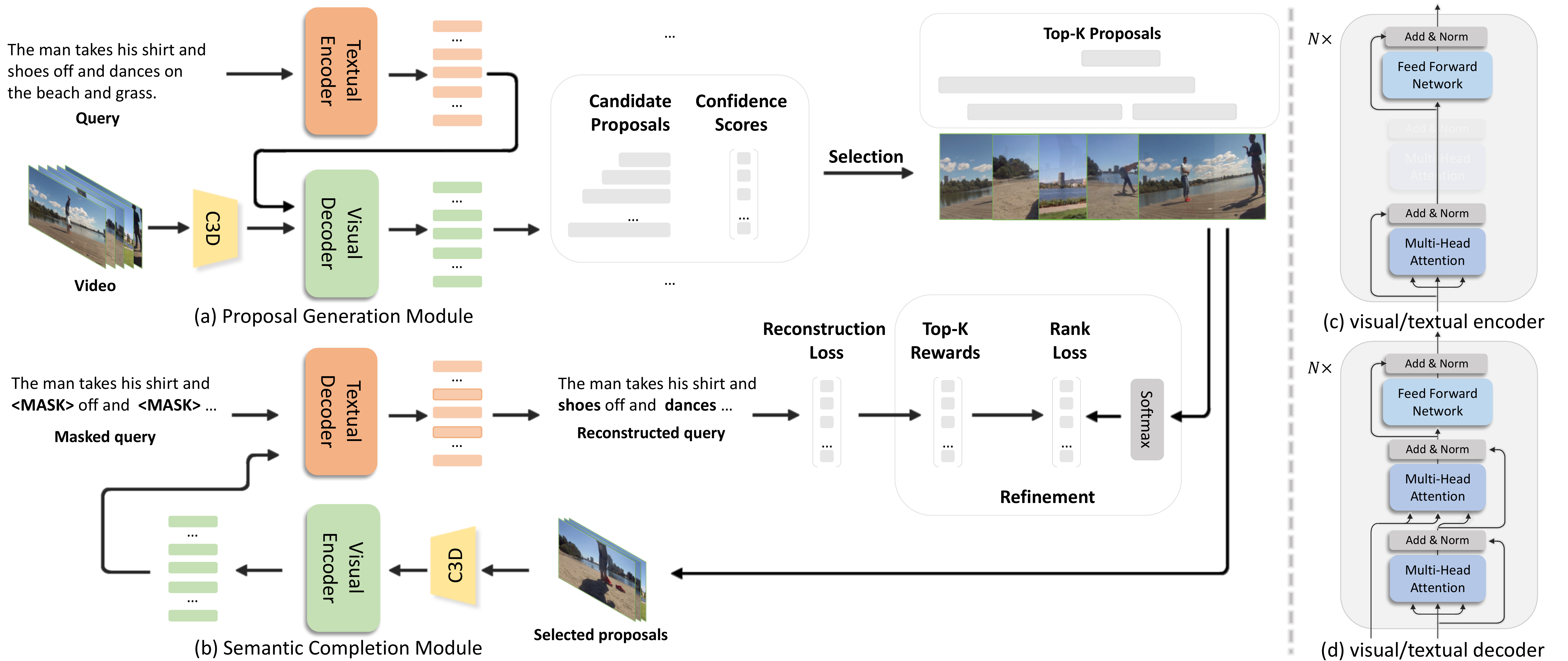} 
    \caption{
        The Framework of our Semantic Completion Network for Video Moment Retrieval.
        (a) The proposal generation module leverages the cross-modal fusion representations of video and
        query to score all the candidate proposals at each time step, 
        and then select the top-K proposals considering both exploitation and exploration.
        (b) The semantic completion module reconstruct the query in which the important words are masked
        according to the visual representations of proposal, compute the rewards based on the reconstruction
        loss, and provide feedbacks to the proposal generation module for scoring refinement.
    }
    \label{fig:framework}
\end{figure*}

\section{Related Work}
In this section, we briefly review some related works on image/video retrieval, 
temporal action detection and video moment retrieval.

\noindent\textbf{Image/Video Retrieval:} Image/Video Retrieval aims to select image/video that is most relevant to 
the queries from a set of candidate images/videos.
The methods in ~\cite{karpathy2015deep,escorcia2016daps,xu2015jointly,otani2016learning} all propose to learn a joint 
visual-semantic space for cross-modal representations. In such space, the similarity of cross-modal representations
reflects the closeness between their original inputs.
In moment retrieval, however, we focus on retrieving a target moment in video based on the given query, 
rather than simply selecting a target image/video from pre-defined candidate sets.

\noindent\textbf{Temporal Action Detection:} Temporal Action Detection aims at identifying the temporal boundary 
as well as the category for each action instance in untrimmed videos. The approaches of action detection can be also
summarized into supervised settings and weakly-supervised settings.
These methods in ~\cite{shou2016temporal,escorcia2016daps,buch2017sst,shou2017cdc,zhao2017temporal} 
are trained in two-stage supervised learning manner, which first generate temporal action proposals 
through a proposal network, and then predict the action category for each proposal through a classification network.
In the weakly-supervised settings, however, only the coarse video-level labels instead of 
the exact temporal boundary is available.
The UntrimmedNet in ~\cite{wang2017untrimmednets} make use of the principle of multiple instance learning
and the generated attention weights to select proposals that most probably contain action instances.
The method presented in ~\cite{nguyen2018weakly} combines temporal class activation maps and class agnostic attentions
for localizing the boundary of action instances.
Further than action detection that is limited to a pre-defined set of categories,
moment retrieval according to natural language query is much more challenging but general.

\noindent\textbf{Video Moment Retrieval:} Video Moment Retrieval is to address the target moment that is semantically aligned with
the given natural language query. 
Prior works~\cite{gao2017tall,hendricks2017localizing,hendricks2018localizing,liu2018cross,chen2018temporally,xu2019multilevel,zhang2019cross,zhang2019man,wang2019language} 
mainly focus on localizing the most relevant moment in a fully-supervised settings.
Among them, methods proposed in~\cite{gao2017tall,hendricks2017localizing,hendricks2018localizing} sample candidate moments 
by sliding windows with various length, and perform coarse fusion to estimate the correlation between 
the queries and moments in a multi-modal space.
Further, the Temporal GroundNet~(TGN)~\cite{chen2018temporally} proposes an interactor to exploit the evolving
fine-grained frame-by-word interactions and simultaneously score a set of candidate moments in one single pass.
The Cross-Modal Interaction Network~(CMIN)~\cite{zhang2019cross} advises a multi-head self-attention mechanism to
capture the long-range dependencies in videos and a syntactic GCN to obtain the fine-grained queries representations.
The Semantic Matching Reinforcement Learning~(SM-RL)~\cite{wang2019language} proposes a recurrent neural network based
reinforcement learning model and introduce mid-level semantic concepts to bridge the semantic gap between visual and semantic
information.

Though those methods achieve good performance, they still suffer from collecting 
a large amount of manually labelled temporal annotations.
Some works~\cite{bojanowski2015weakly,duan2018weakly,mithun2019weakly} also study this task 
in a weakly-supervised setting.
The method proposed in ~\cite{bojanowski2015weakly} consider the task of aligning a video with a set of temporal ordered
sentences, in which temporal ordering can be seen as additional constraint and supervision.
The method proposed in ~\cite{duan2018weakly} decomposes the problem of 
weakly-supervised dense event captioning in videos~(WS-DEC) into a cycle of dual problems: 
caption generation and moment retrieval and explores the one-to-one correspondence 
between the temporal segment and event caption, 
and has a complex training pipeline such as pre-training and alternating training.
The Text-Guided Attention~(TGA)~\cite{mithun2019weakly} proposes to learn a joint visual-semantic representations
and utilizes the attention score as the alignment between video frames and query.

\section{Approach}

\subsection{Problem Formulation}
In this paper, we consider the task of Video Moment Retrieval in a weakly-supervised setting.
Given an untrimmed video ${\bf v} = \{{\bf v}_i\}_{i=1}^{n_v}$ 
where $n_v$ is the frame number of the video and ${\bf v}_i$ is the $i$-th feature vector,
and a corresponding query ${\bf q} = \{{\bf q}_i\}_{i=1}^{n_q}$
where $n_q$ is the word number of the query and ${\bf q}_i$ is the $i$-th feature vector,
aims to localize the most relevant moment ${\hat \tau} = ({\hat s},~{\hat e})$ during inference,
where ${\hat s},~{\hat e}$ are the indices of start frame and end frame respectively.

\subsection{Proposal Generation Module}
In this section, we introduce the proposal generation module.
As mentioned above, the attention weights usually focus on the most discriminative but small regions,
and thus fails to cover the entire temporal extent of target moment.
As Figure~\ref{fig:framework}(a) shows, instead,
this module scores the candidate proposals according to the cross-modal representations of video
and query.
Moreover, further than these methods in ~\cite{gao2017tall,hendricks2018localizing} that 
handle different proposals separately in a sliding window fashion, our method scores all the 
candidate moments in a single pass, which makes full use of the context information.

In detail, the feature vector ${\bf q}_i$ of each word
can be extracted using a pre-trained word2vec embedding.
Then we develop a textual encoder ${\bf Enc}_q$ to obtain 
the textual representations for the query ${\bf q}$.
After that, we input the textual representations and
the video features ${\bf v}^i$ to the visual decoder ${\bf Dec}_v$ to obtain
the final cross-modal representations ${\bf c} = \{{\bf c}_i\}_{i=1}^{n_v}$ of video and query, given by
\begin{eqnarray}
{\bf c} =  {\bf Dec}_v({\bf v},~{\bf Enc}_q({\bf q})),
\end{eqnarray}

To generate confidence score in a single pass, we first pre-define a set of candidate proposals
at each time step, 
denoted by $C_t = \{(t - r_k * n_v,~t)\}_{k=1}^{n_k}$, 
where $t - r_k * n_v,~t$ are the start and end boundaries of the $k$-th candidate proposal
at the $t$-th time step, $r_k \in (0, 1)$ is the $k$-th ratio and $n_k$ is the number of candidate proposals.
Note that $r_k$ is a fix ratio for each time step.
Then based on the cross-modal representations ${\bf c}$,
we can simultaneously give the confidence scores for 
these proposals at all time steps by a fully connected layer 
with sigmoid nonlinearity, denoted by
\begin{eqnarray}
    SC_t = \sigma({\bf W}_s{\bf c}_t + {\bf b}_s),
\end{eqnarray}
where $SC_t \in \mathcal{R}^{n_k}$ represents the vector of confidence scores for the $n_k$ candidate proposals
at the $t$-th time step.

Given the candidate proposals $\{C_t\}_{t=1}^{n_v}$, 
we apply the selection algorithm that considers both exploitation and exploration
to select the top-K proposals $G = \{G^{k}\}_{k=1}^{K}$ 
and give the corresponding confidence scores $S = \{S^{k}\}_{k=1}^{K}$,
where $G^{k} = (s_k,~e_k)$ represents the $k$-th proposal in top-K proposals and
$S^{k}$ is its confidence score.
Concretely, we rank the proposals according to their corresponding confidence scores.
At each step, we choose a proposal randomly with a possibility of $p$ or choose
the proposal with the highest score with a possibility of $1 - p$, and use Non Maximum Suppression~(NMS) to
remove those proposals that have high overlap with the chosen one.
We define the sampling possibility $p$ with a decay function dependent on 
the times of parameter updates $n_{update}$, given by
\begin{eqnarray}
    p = \lambda_1 * {\rm exp}(-n_{update} / \lambda_2),
\end{eqnarray}
where $\lambda_1$, $\lambda_2$ are the hyper-parameters to control the decay rate.
As the training proceeds, the possibility of choosing next proposal randomly
decreases gradually.

\subsection{Semantic Completion Module}
In this section, we introduce the semantic completion module to 
measure the semantic similarity between proposals and query,
compute rewards
and provide feedbacks to previous module for scoring refinement.
As shown in Figure~\ref{fig:framework}(b), 
the important words~(e.g. noun, verb) are masked and predicted
according to the given visual context.
The most semantically matching proposal can 
provide enough useful information to predict the key words
and also contains less noise.

First, we extract video features for the $k$-th proposal $G^{k} = (s_k,~e_k)$,
denoted by ${\bf \hat v}^k = \{{\bf v}_{i}\}_{i=s_k}^{e_k}$, and obtain 
the visual representations 
through the visual encoder ${\bf Enc}_v$.
We denote the original words sequence as ${\bf w} = \{{\bf w}_i\}_{i=1}^{n_q}$,
where ${\bf w}_i$ is the $i$-th word of the query.
Then given the words sequence ${\bf w} $ and a set of masked position $\mathcal{X}$, 
we denote ${\bf \hat w}$ as a modified version of ${\bf w}$ 
where those words ${\bf w}_i$, $i \in \mathcal{X}$ are replaced by a special symbol.
We can extract word features for ${\bf \hat w}$, denoted as ${\bf \hat q} = \{{\bf \hat q}_i\}_{i=1}^{n_q}$.
Next, through a bi-directional textual decoder ${\bf Dec}_q$, we can obtain the final cross-modal semantic 
representations ${\bf f}^{k} = \{{\bf f}^{k}_i\}_{i=1}^{n_q}$ for the proposal $G^{k}$, given by
\begin{eqnarray}
    {\bf f}^{k} = {\bf Dec}_q({\bf \hat q},~{\bf Enc}_v({\bf \hat v}^{k})),
\end{eqnarray}

To predict the masked words, we can compute the energy distribution ${\bf e}^{k} = \{{\bf e}^{k}_i\}_{i=1}^{n_q}$
on the vocabulary by a fully connected layer, denoted by
\begin{eqnarray}
    {\bf e}^{k}_i = {\bf W}_v{\bf f}^{k}_i + {\bf b}_v,
\end{eqnarray}
where ${\bf e}^{k}_i \in \mathcal{R}^{n_w}$ is the energy distribution at the $i$-th time step, 
$n_w$ is the number of words in the vocabulary.

\subsection{Training of Semantic Completion Network}
In this section, we describe the loss function we optimize to train the Semantic Completion Network.

\noindent\textbf{Reconstruction Loss.} 
With the energy distribution ${\bf e}^{k}$ for the proposal $G^k$,
we first adopt a reconstruction loss to train the semantic
completion module and make it able to extract key information from the visual context
to predict the masked words. 
Formally, we then compute the negative log-likehood of each masked word and add them up, 
denoted by
\begin{eqnarray}
    \mathcal{L}_{rec}^{k} 
    = -\sum_{i=1}^{n_q-1}{\rm log}~p({\bf w}_{i+1}|{\bf \hat w}_{1:i}, {\bf \hat v}^{k}) \\ 
    = -\sum_{i=1}^{n_q-1}{\rm log}~p({\bf w}_{i+1}|{\bf e}^{k}_i),
\end{eqnarray}
where $\mathcal{L}_{rec}^{k}$ represents the reconstruction loss based on the visual context of
the proposal $G^{k}$.

\noindent\textbf{Rank Loss.} As Figure~\ref{fig:framework} shows, 
in order to correct the confidence scores given by the proposal generation module, 
we further apply a rank loss to train this module. 
Note that we correct the confidence scores based on reward rather than one-hot label.
Specifically,
we define the reward $R^{k}$ for the proposal $G^{k}$ 
with a reward function to encourage
proposals with lower reconstruction loss.
The reward is reduced from one to zero in steps of $1~/~(K - 1)$.

Then the strategy of policy gradient is used to correct the scores.
Note that the confidence scores are normalized by a $softmax$ layer, which is 
an extremely important operation to highlight the semantically matching proposals
and weaken the mismatched ones.
The rank loss $\mathcal{L}_{ran}^{k}$ for the proposal $G^{k}$ is computed by
\begin{eqnarray}
    \mathcal{L}_{ran}^{k} = -{R^{k}~{\rm log}~(
        \frac{{\rm exp}(S^{k})}{\sum_{i=1}^{K}{\rm exp}(S^{i})})},
\end{eqnarray}

\noindent\textbf{Multi-Task Loss.} With the reconstruction loss and the rank loss for each proposal, we average losses over all proposals and
compute a multi-task loss to train the semantic complete network in an end-to-end manner, denoted by
\begin{eqnarray}
    \mathcal{L} = \frac{1}{K}\sum_{k=1}^K({\mathcal{L}_{rec}^{k}} + \beta\mathcal{L}_{ran}^{k}),
\end{eqnarray}
where $\beta$ is a hyper-parameter to control the balance of two losses.

\subsection{Network Design}
In this section, we introduce the details of the semantic completion network, including
the components of visual/textual encoder and visual/textual decoder.

\noindent\textbf{Encoder and Decoder.} It has been indicated in ~\cite{tang2018self} 
that Transformer~\cite{vaswani2017attention} is a strong feature extractor.
In this paper, we build our visual/textual encoder and visual/textual decoder based on 
the bi-directional Transformer, as Figure~\ref{fig:framework}(c)(d) shows.
The encoder/decoder is composed of a stack of layer that contains
the multi-head attention sub-layer and the fully connected feed-forward network.

\noindent\textbf{Parameter Sharing.} We share the parameters between 
the visual/textual encoder and visual/textual decoder.
As Figure~\ref{fig:framework}(c)(d) shows,
an encoder 
can be also regarded as a decoder
without computing attention from another input of different modality.
Parameter sharing greatly reduces the number of parameters and save memory.
This is also a kind of model-level dual learning~\cite{xia2018model} sharing parameters across tasks,
which promotes knowledge sharing.

\section{Experiments}

\subsection{Datasets}
We perform experiments on two public datasets for video moment retrieval
to evaluate the effectiveness of our SCN method.

\noindent\textbf{ActivityCaptions.} 
The ActivityCaptions~\cite{caba2015activitynet} dataset is originally developed for human activity understanding.
This dataset contains 20,000 various untrimmed videos 
and each video includes multiple natural language descriptions with temporal annotations. 
The released ActivityCaptions dataset comprise 
17,031 description-moment pairs for training.
Since the caption annotations of test data of ActivityCaptions are not publically available, 
we take the val\_1 as the validation set and val\_2 as test data. 
The average length of the description, also regarded as query in moment retrieval,
is 13.16 words, and the average duration of the video is 117.74 seconds.

\noindent\textbf{Charades-STA.} 
The Charades-STA dataset is released in ~\cite{gao2017tall} for moment retrieval,
and comprises 12,408 description-moment pairs for training, and 3,720 for testing.
The average length of the query is 8.6 words, 
and the average duration of the video is 29.8 seconds.
The Charades dataset, originally introduced in ~\cite{sigurdssonhollywood},
contains only temporal activity annotation and multiple video-level descriptions for each video. 
The authors of ~\cite{gao2017tall} design a semi-automatic way 
to generate sentence temporal annotations.
First, the video-level descriptions from the original dataset 
were split into sub-sentences. 
Then, by matching keywords for activity categories,
these sub-sentences are aligned with moments in videos. 
The rule-based annotations are ultimately verified by humans.

\subsection{Evaluation Metric}
To evaluate the performance of our SCN method and baselines,
we adopt the evaluation metric proposed by ~\cite{gao2017tall}
to compute ``R@n,~IoU=m''.
Specifically, we compute the percentage of at least one of the top-n
predicted moments having Intersection over Union~(IoU) larger than m,
denoted by $R(n,m) = \frac{1}{n_t}\sum_{i=1}^{n_t}{r(n,m,q_i)}$,
where $q_i$ is the $i$-th query, $n_t$ is the number of testing query,
${r(n,m,q_i)}$ is 1 only if the top-n returned moments about $q_i$ 
contains at least one that has a temporal IoU $>$ m
and $R(n,m)$ is the overall performance.

\subsection{Implementation Details}
\textbf{Data Preprocessing.} For each video, we pre-extract visual frame-based features 
by a publicly available pre-trained 3D-ConvNet model
which has a temporal resolution of 16 frames.
This network was not fine-tuned on our data. 
We reduce the dimensionality of the activations from the second fully-connected layer (fc7) 
of the network from 4096 to 500 dimensions using PCA. 
The C3D features were extracted every 8 frames.
The maximum number of frame is set to 200.

For each description, we split it into words using NLTK and extract word embeddings using
the pretrained Glove~\cite{pennington2014glove} word2vec for each word token.
The maximum description length is set to 20.
We also keep the most common $n_w$ words in training set, resulting in a vocabulary size of 8,000
for ActivityCaptions and 1,111 for Charades-STA.

\noindent\textbf{Model Settings.} 
At each time step of video, we score $n_k$ candidate proposals of multiple scales.
We set $n_k$ to 6 with ratios of ~[0.167, 0.333, 0.500, 0.667, 0.834, 1.0] for ActivityCaptions, 
and to 4 with ratios of ~[0.167, 0.250, 0.333, 0.500] for Charades-STA.
We then set the decay hyper-parameter $\lambda_1$ to 0.5,
$\lambda_2$ to 2000,
the number of selected proposals $K$ to 4,
the balance hyper-parameter $\beta$ to 0.1.
Also, we mask one-third of words in a sentence and replace with a special token
for semantic completion.
Note that noun and verb are more likely to be masked.
Moreover, for TransformerEncoder as well as TransformerDecoder, 
the dimension of hidden state is set to 256
and the number of layers is set to 3.
During training,
we adopt the Adam optimizer with learning rate 0.0002 to
minimize the multi-task loss. 
The learning rate increases linearly to the maximum with a warm-up step of 400
and then decreases itself based on
the number of updates~\cite{vaswani2017attention}.

\subsection{Compared Methods} 
\textbf{Random.} We simply select a candidate moment randomly.

\noindent\textbf{VSA-RNN and VSA-STV.}~\cite{gao2017tall} This two methods both simply project 
the visual feature of all candidate proposals and the textual feature of the query into a common space,
and computes the confidence scores based on cosine similarity.

\noindent\textbf{CTRL.}~\cite{gao2017tall} The CTRL method introduces a cross-modal temporal
localizer to estimate the alignment scores and uses clip location regression to further adjust the
the boundary.

\noindent\textbf{QSPN.}~\cite{xu2019multilevel} The QSPN method devises a multilevel approach
for integrating vision and language features using attention mechanisms, and also leverages
video captioning as an auxiliary task.

\noindent\textbf{WS-DEC.}~\cite{duan2018weakly} The WS-DEC method decomposes the problem of 
weakly-supervised dense event captioning in videos into a cycle of dual problems: 
caption generation and moment retrieval, and explores the one-to-one correspondence 
between the temporal segment and event caption.

\noindent\textbf{TGA.}~\cite{mithun2019weakly} The TGA method proposes a weakly-supervised 
joint visual-semantic embedding framework for moment retrieval, and utilizes the latent
alignment for localization during inference.

\subsection{Quantitative Results and Analysis}
The overall performance results of our SCN and baselines on ActivityCaptions and Charades-STA datasets
are presented in Table~\ref{table:activityres} and Table~\ref{table:charadesres} respectively.
We consider the evaluation metric ``R@n,~IoU=m'', 
where $n \in \{1,5\}, m \in \{0.1,0.3,0.5\}$ for ActivityCaptions, 
and $n \in \{1,5\}, m \in \{0.3,0.5,0.7\}$ for Charades-STA.
By observing the evaluation results, we can discover some facts:

\begin{itemize}
    \item Compared with Random method, 
    the overall performance results of SCN have a huge improvements on
    both two datasets, which demonstrates that optimizing the multi-task loss
    instead of explicitly optimizing the localization loss can reach the goal
    of predicting the target moment and also indicates the feasibility of our SCN method.
    \item As the results show, the proposed SCN method outperforms 
    the supervised visual-embedding approaches VSA-RNN and VSA-STV significantly, 
    and obtains results comparable to the other fully-supervised methods
    on two datasets,
    indicating that even without the full annotations of temporal boundary,
    our SCN method can still 
    effectively exploit the alignment relationship between video and query 
    and find the most semantically relevant moment.
    \item The coarse methods VSA-RNN and VSA-STV achieve the worst performance on two datasets,
    even compared with the weakly-supervised SCN method,
    demonstrating the key role of visual and textual modeling in moment retrieval and indicating
    the limitation of learning a common visual-semantic space in high-quality retrieval.
    \item Also, compared with the weakly-supervised methods WS-DEC and TGA, 
    our method achieves tremendous improvements on both ActivityCaptions and
    Charades-STA datasets.
    These results verify the effectiveness of the proposal generation module,
    the semantic completion module, the algorithm of proposals selection and
    the multi-task loss.
\end{itemize}

\begin{table}[t]
    \centering
    \caption{Performance Evaluation Results on the ActivityCaptions Dataset 
    ($n \in \{1,5\}$ and  $m \in \{0.1,0.3,0.5\}$). }
    \label{table:activityres}
    \scalebox{0.7}{
        \begin{tabular}{c|ccc|ccc}
            \hline
            \hline
            \multirow{2}{*}{Method} & \multicolumn{3}{c|}{R@1} & \multicolumn{3}{c}{R@5} \\
                &  IoU=0.1&   IoU=0.3 &  IoU=0.5 & IoU=0.1&   IoU=0.3 &  IoU=0.5 \\
            \hline
                Random&38.23&18.64&7.63
                      &75.74&52.78&29.49\\
            \hline
                VSA-RNN&-&39.28&23.43
                       &-&70.84&55.52\\
                VSA-STV&-&41.71&24.01
                       &-&71.05&56.62\\
                CTRL&-&47.43&29.01
                    &-&75.32&59.17\\
                QSPN&-&52.12&33.26
                    &-&77.72&62.39\\
            \hline
                WS-DEC&62.71&41.98&23.34
                      &-&-&-\\
                SCN&{\bf 71.48}&{\bf 47.23}&{\bf 29.22}
                   &{\bf 90.88}&{\bf 71.45}&{\bf 55.69}\\
            \hline
        \end{tabular}
    }
\end{table}

\begin{table}[t]
    \centering
    \caption{Performance Evaluation Results on the Charades-STA Dataset 
    ($n \in \{1,5\}$ and  $m \in \{0.3,0.5,0.7\}$). }
    \label{table:charadesres}
    \scalebox{0.7}{
        \begin{tabular}{c|ccc|ccc}
            \hline
            \hline
            \multirow{2}{*}{Method} & \multicolumn{3}{c|}{R@1} & \multicolumn{3}{c}{R@5} \\
                &  IoU=0.3&   IoU=0.5 &  IoU=0.7 & IoU=0.3&   IoU=0.5 &  IoU=0.7\\
            \hline
                Random&20.12&8.61&3.39
                      &68.42&37.57&14.98\\
            \hline
                VSA-RNN&-&10.50&4.32
                       &-&48.43&20.21\\
                VSA-STV&-&16.91&5.81
                       &-&53.89&23.58\\
                CTRL&-&23.63&8.89
                    &-&58.92&29.52\\
                QSPN&54.70&35.60&15.80
                    &95.60&79.40&45.40\\
            \hline
                TGA&32.14&19.94&8.84
                   &86.58&65.52&33.51\\
                SCN&{\bf 42.96}&{\bf 23.58}&{\bf 9.97}
                   &{\bf 95.56}&{\bf 71.80}&{\bf 38.87}\\
            \hline
        \end{tabular}
    }
\end{table}

\begin{figure}[t]
    \centering
    \subfigure[ActivityCaptions]{
        \includegraphics[width=0.45\columnwidth]{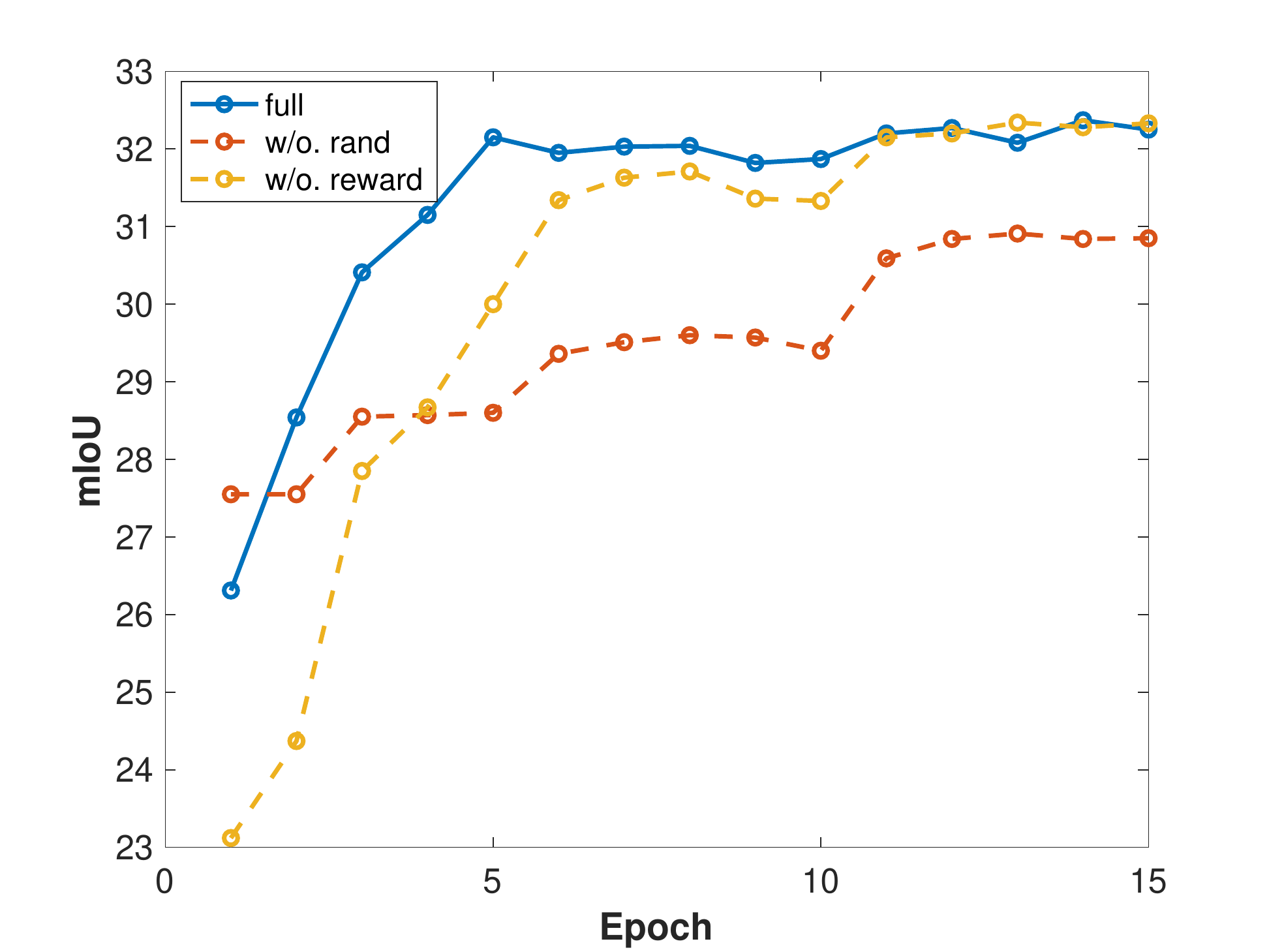}
    }
    \subfigure[Charades-STA]{
        \includegraphics[width=0.45\columnwidth]{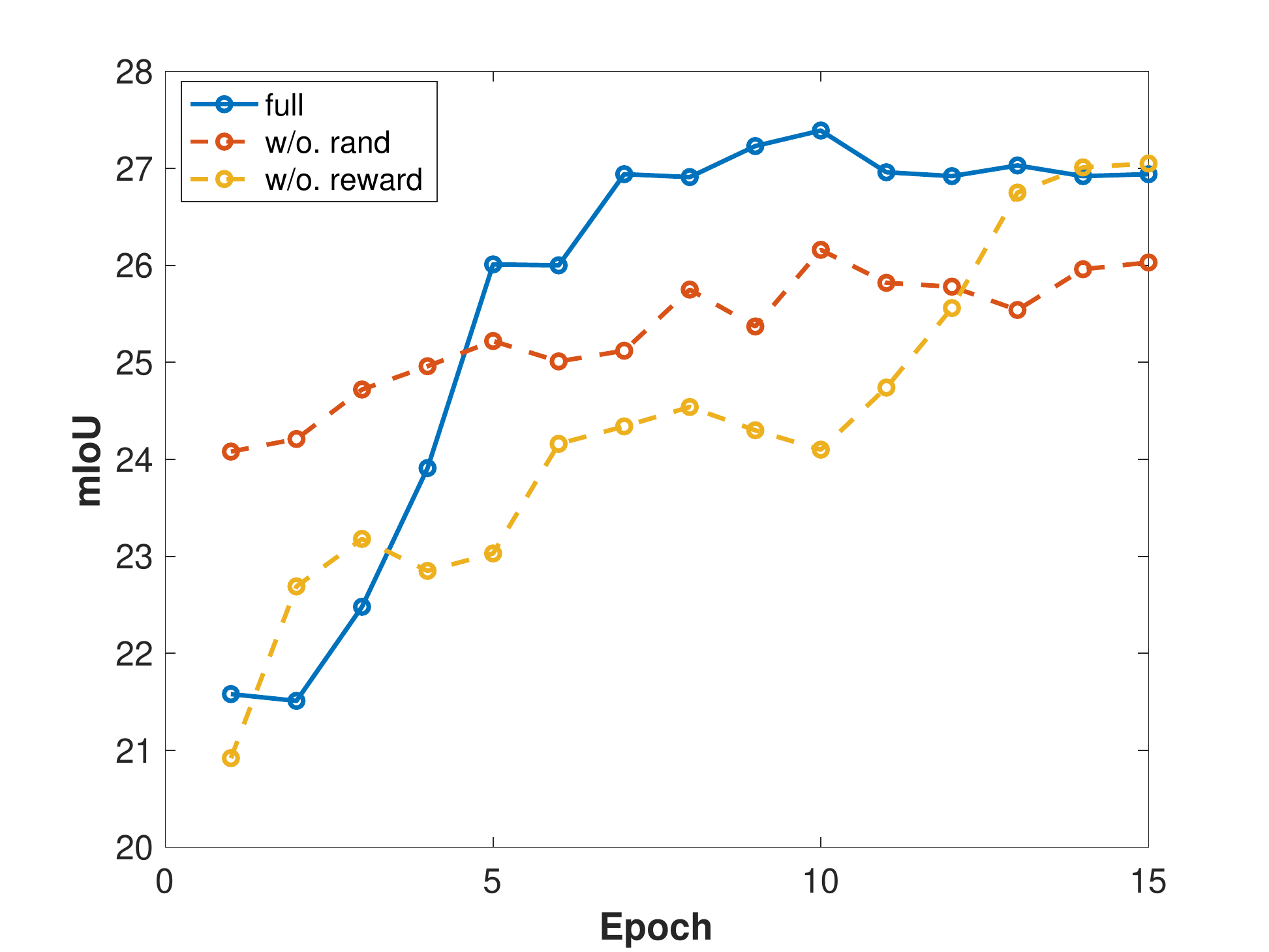}
    }
    \caption{Training Process of Different Models 
    on ActivityCaptions and Charades-STA Datasets}
    \label{fig:ab_compare1} 
\end{figure}

\begin{figure}[t]
    \centering
    \subfigure[ActivityCaptions]{
        \includegraphics[width=0.45\columnwidth]{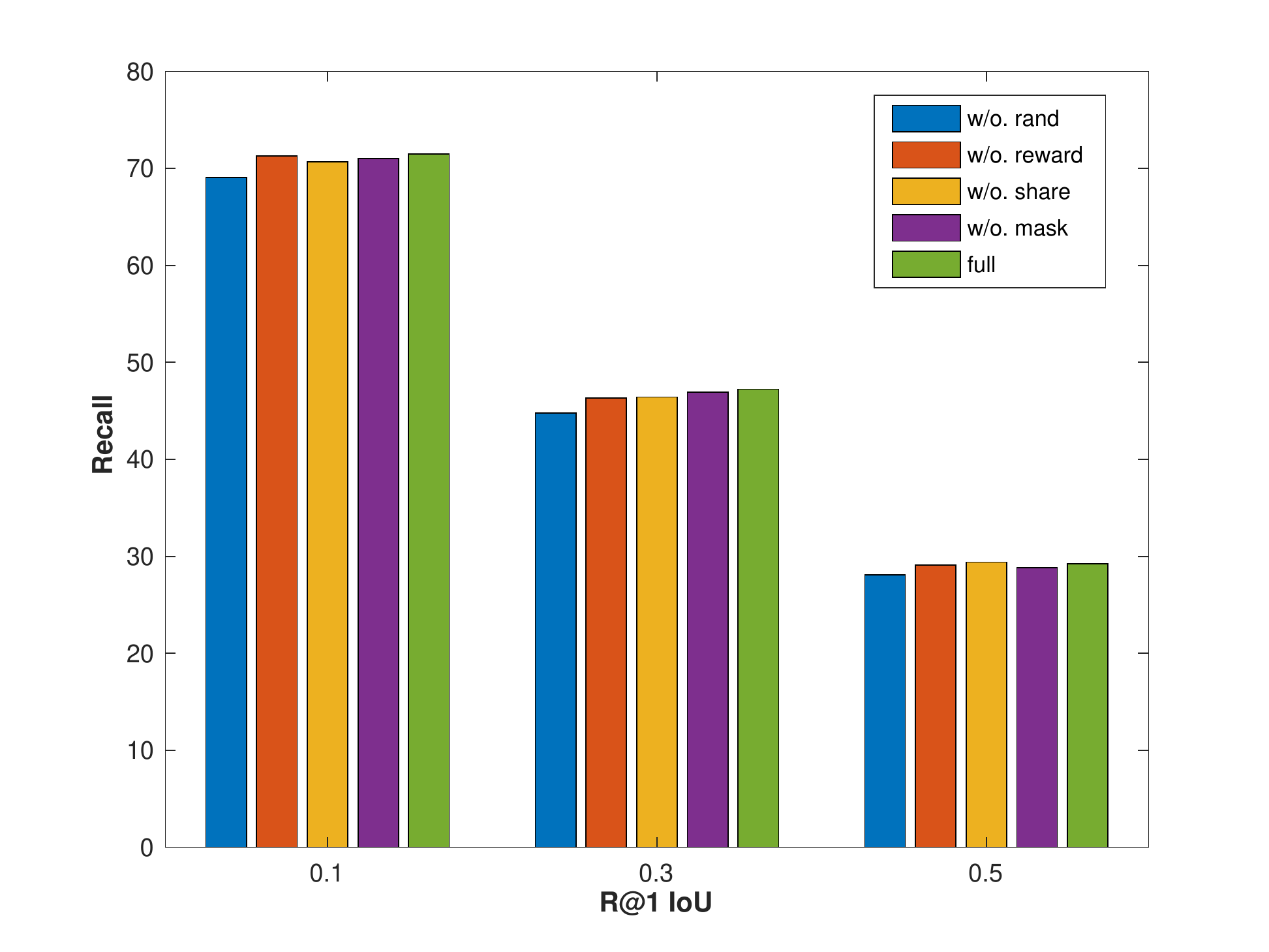}
    }
    \subfigure[Charades-STA]{
        \includegraphics[width=0.45\columnwidth]{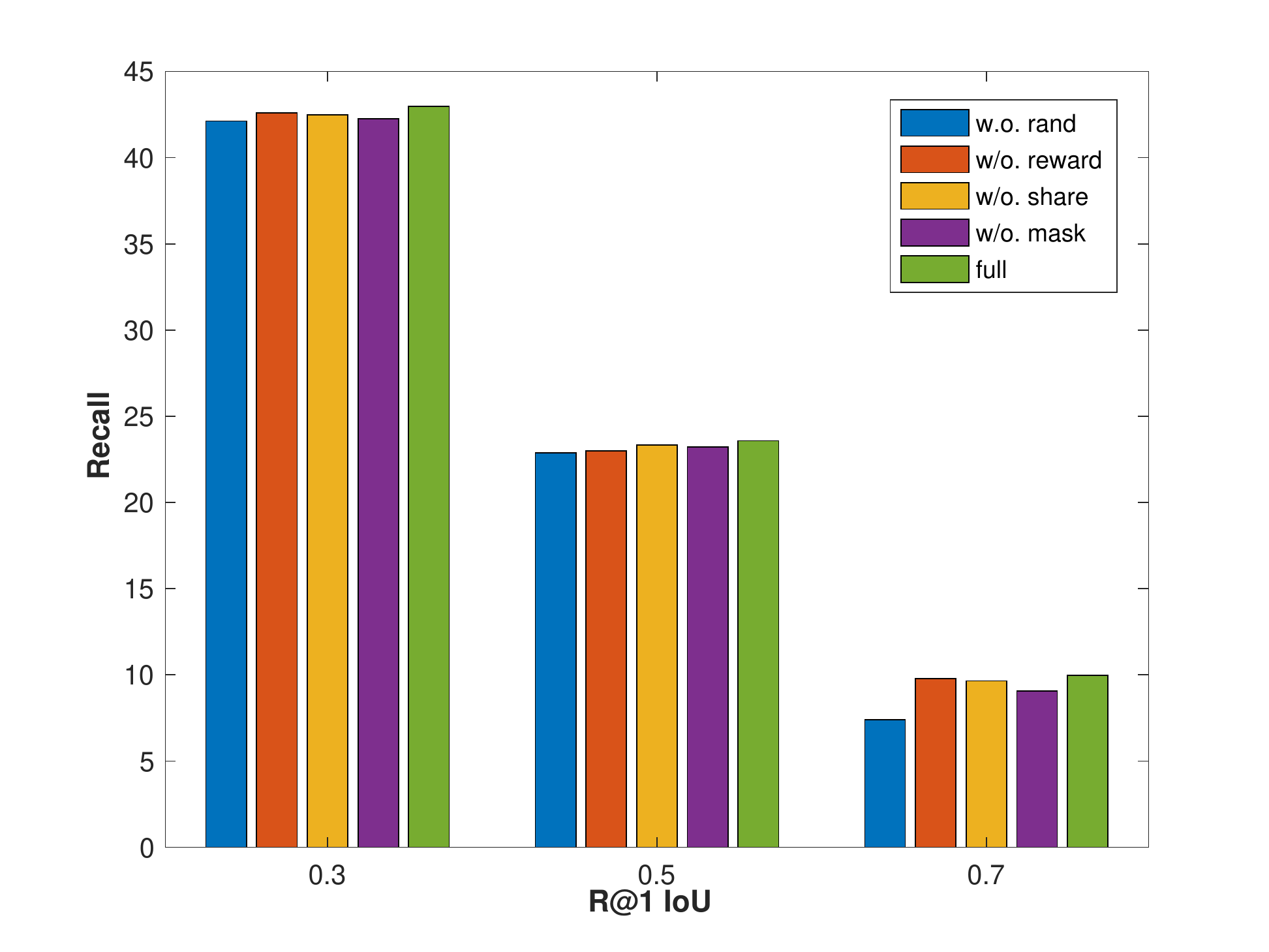}
    }
    \caption{Evaluation Results of Different Models 
    on ActivityCaptions and Charades-STA Datasets}
    \label{fig:ab_compare2} 
\end{figure}


\begin{figure*}[t]
    \centering
    \includegraphics[width=1.0\columnwidth]{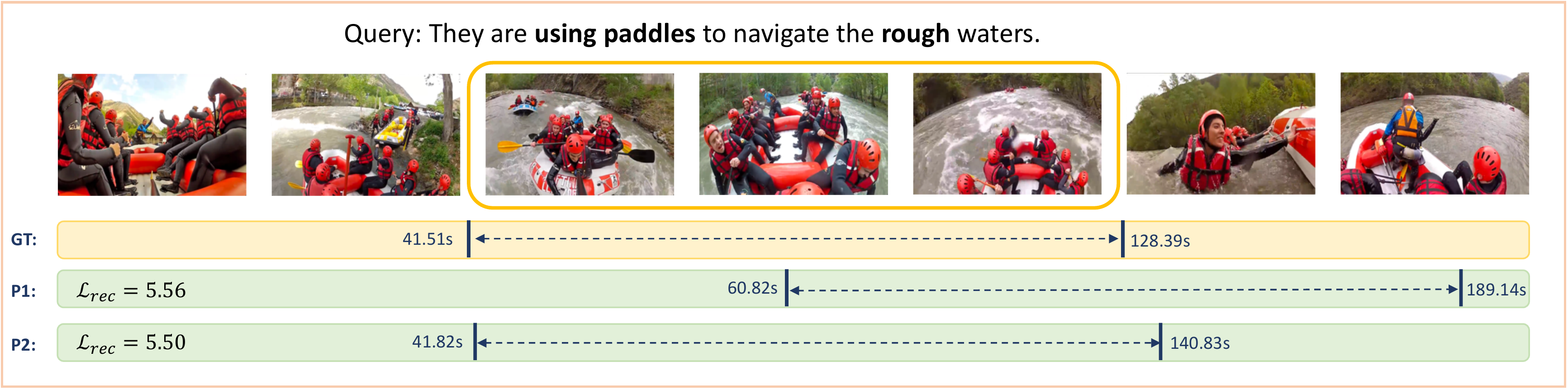} 
    \includegraphics[width=1.0\columnwidth]{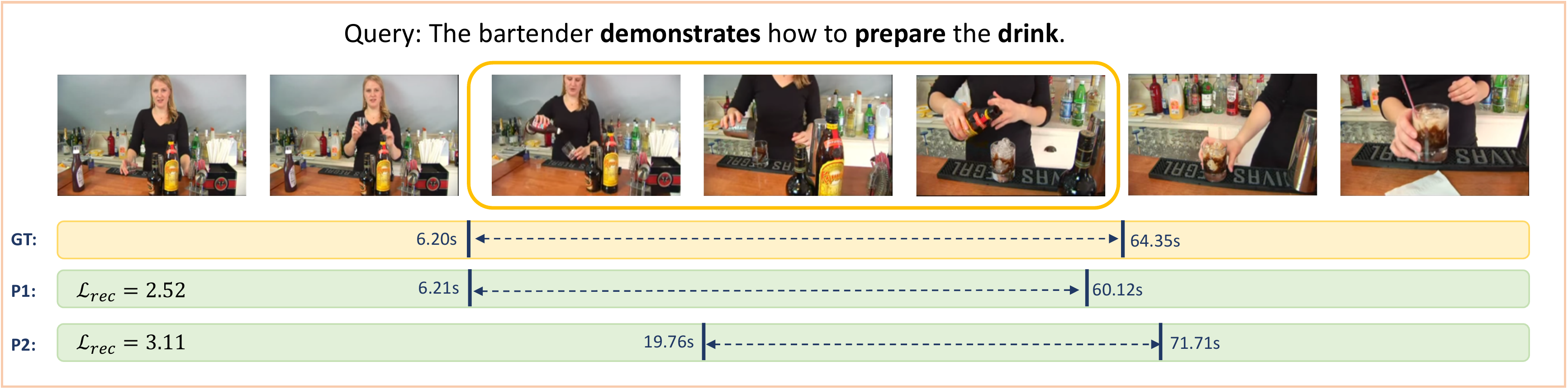} 
    \includegraphics[width=1.0\columnwidth]{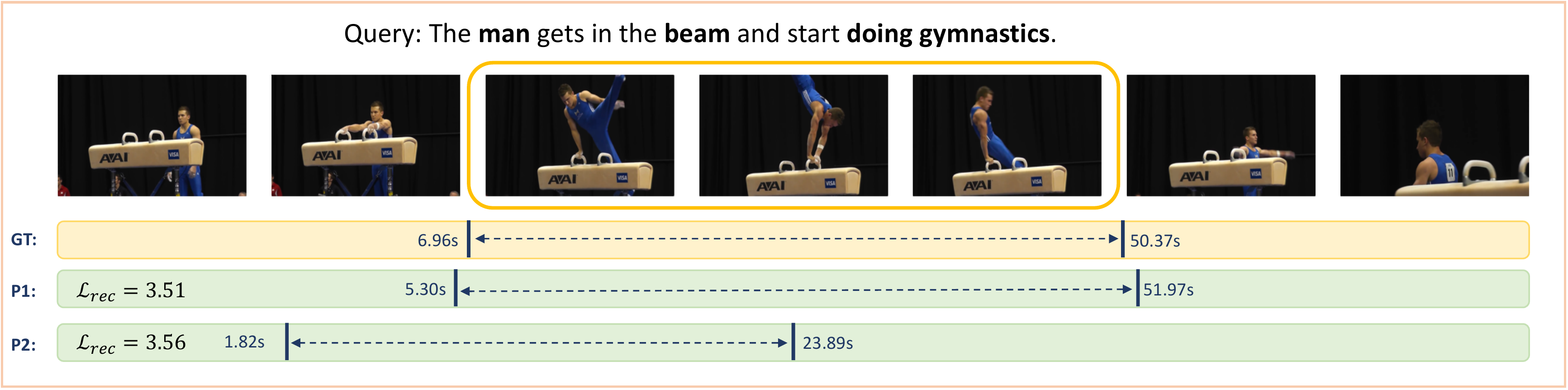} 
    \includegraphics[width=1.0\columnwidth]{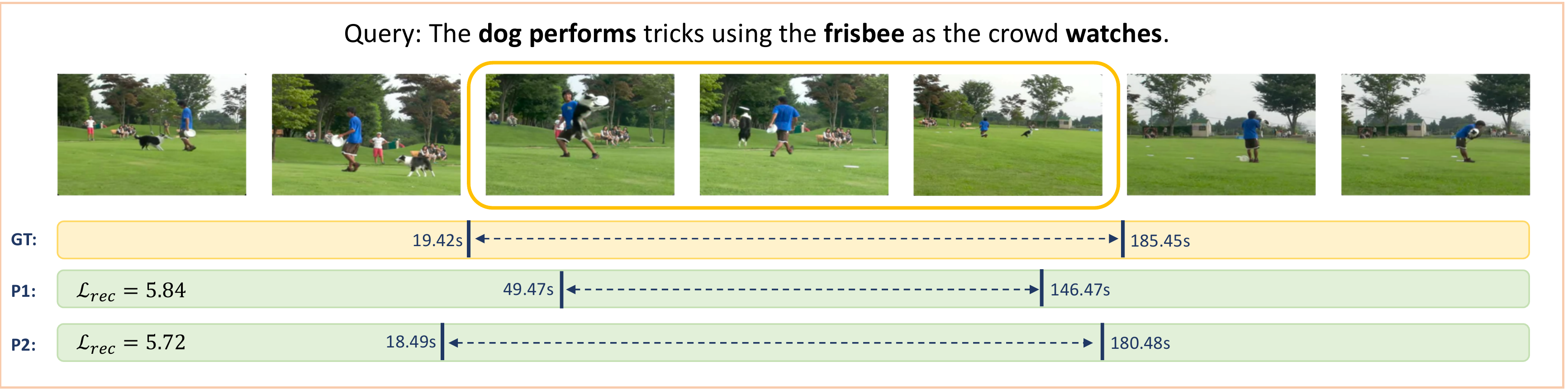} 
    \caption{Qualitative Examples on the ActivityCaptions dataset}
    \label{fig:example_act}
\end{figure*}

\begin{figure*}[!ht]
    \centering
    \includegraphics[width=1.0\columnwidth]{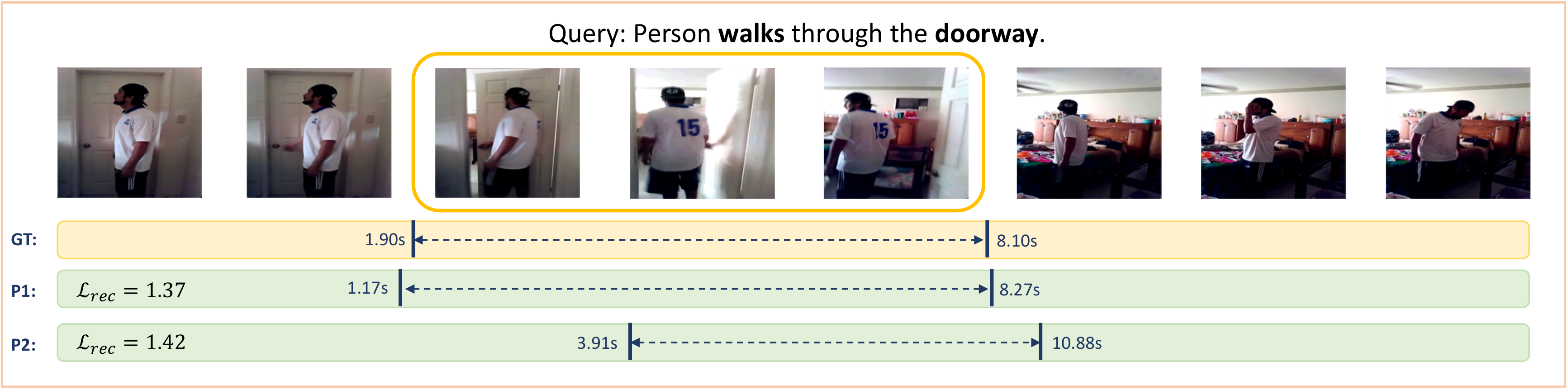} 
    \includegraphics[width=1.0\columnwidth]{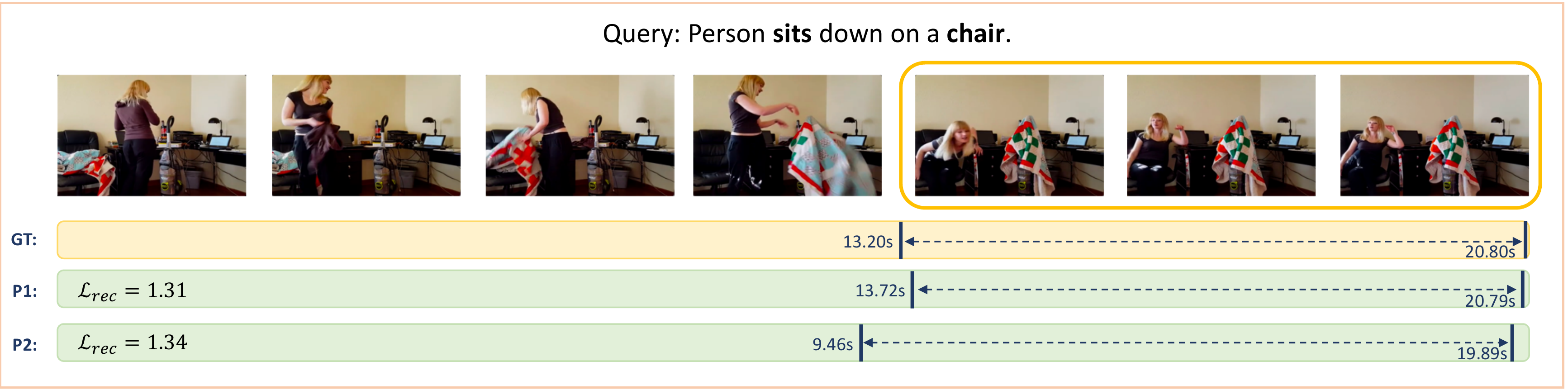} 
    \caption{Qualitative Examples on the Charades-STA dataset}
    \label{fig:example_cha}
\end{figure*}

\subsection{Ablation Study}
To prove the validity of different parts of our method, 
we simplify the algorithm to generate different ablation models as follows:
\begin{itemize}
    \item \textbf{SCN(w/o. rand).} During proposals selection, 
    we assign the sample possibility $p$ to zero,
    which means we select next proposal 
    completely based on the confidence scores without random selection
    at each step.
    \item \textbf{SCN(w/o. reward).} With feedbacks given by the semantic completion module, 
    we modify the rank loss by using one-hot label instead of computing rewards for scoring refinement.
    Concretely, we simply assign a reward of one to the best proposal, and zero to the other ones. 
    The rank loss is equivalent to the cross entropy loss.
    \item \textbf{SCN(w/o. mask).} To validate the effectiveness of the semantic completion module 
    and the reconstruction loss, we replace this module with a ordinary captioning generator~\cite{duan2018weakly}
    without masking words.
    \item \textbf{SCN(w/o. share).} Instead of parameter sharing between the proposal generation module 
    and the semantic completion module, we use two separate sets of parameters for this two modules.
\end{itemize}
The training process of different models on ActivityCaptions and Charades-STA
is presented in Figure~\ref{fig:ab_compare1}.
By analyzing the results, we can find some interesting points:
\begin{itemize}
    \item The simplified models SCN(w/o. rand) and SCN(w/o. reward) still 
    achieve results comparable to the fully-supervised methods and outperform
    the existing weakly-supervised methods, which further demonstrates 
    the effectiveness of our framework 
    including proposal generation and selection, 
    semantic completion for semantic similarity estimation and scoring refinement.
    \item The SCN(full) achieves better results than the SCN(w/o. rand) and 
    the evaluation results of SCN(w/o. rand) grows more gently as the model converges,
    which proves the ability of the random selection to find potentially good proposals
    during proposals selection. 
    When the model has not converged, selecting proposals randomly provide opportunities for 
    those potentially good proposals and speed up training.
    \item The SCN(full) achieves the best results faster than SCN(w/o. reward), which demonstrates
    the effectiveness of employing rewards as feedbacks to train the proposal generation module. 
    In the early of training stage, the semantic module can't provide accurate feedbacks 
    but the one-hot label force the previous module to accept only one proposal 
    and reject other proposals that are actually reasonable.
    \item As shown in ~\ref{fig:ab_compare2}, the SCN(full) achieves better
    results than the SCN(w/o. mask), which indicating the effectiveness of 
    the semantic completion module and the masking operation. 
    By masking the important words of the query, we forces the decoder to absorb the cross-modal
    visual information in the decoder side.
    \item The SCN(full) also perform a bit better than the SCN(w/o. share), demonstrating
    the effectiveness of parameter sharing to promote knowledge sharing. 
    Also, The amount of parameters is greatly reduced by parameter sharing.
\end{itemize}

\subsection{Qualitative Results}
To qualitatively validate the performance of our SCN method,
several examples of video moment retrieval from ActivityCaptions and Charades-STA
are provided in Figure~\ref{fig:example_act} and Figure~\ref{fig:example_cha} respectively.
Each example provide the ground truth of temporal boundaries, 
the first two proposals with the highest confidence score given 
by the proposal generation module. 
The bold words in the sentence are considered as the important words 
associated with the video context, and are masked for semantic completion.
The corresponding reconstruction loss $\mathcal{L}_{rec}$ is also computed 
and presented in each example.

It can be observed that both the first two proposals with the highest confidence score
cover the most discriminative video contents relevant to the query, 
which qualitatively verify that the proposal generation module
can locate those semantically important proposals,
and the rank loss is helpful for scoring refinement during training.
Additionally, the proposal with higher IoU has lower reconstruction loss, 
also indicating the proposal that is more semantically matching with the query 
can be recognized by the semantic completion module.
Therefore, due to the effectiveness of two submodules and the training algorithm, our method is
successful in localizing the moment that has high IoU with the target moment.

\section{Conclusion}
In this paper, we study the task of video moment retrieval from the perspective 
of weak-supervised learning 
without manually-labelled temporal boundaries of start time and end time, 
which makes this task more realistic but more challenging.
We propose a novel semantic completion network~(SCN) 
including the proposal generation module to score all candidate proposals in a single pass,  
an efficient algorithm for proposals selection considering both exploitation and exploration,
the semantic completion module for semantic similarity estimation
and a multi-task loss for training.
The experiments on the ActivityCaptions and Charades-STA datasets also demonstrate
the effectiveness of our method to exploit the alignment relationship between
video and query, and the efficiency of the proposal selection algorithm
and the rank loss.

\section{Acknowledgements}
This work is supported by the National Natural Science Foundation of China under
Grant No.61602405, No.U1611461, No.61751209 and No.61836002,
China Knowledge Centre for Engineering Sciences and Technology,
and Alibaba Innovative Research.

\bibliography{ref.bib}

\begin{thebibliography}{}

\bibitem[\protect\citeauthoryear{Bojanowski \bgroup et al\mbox.\egroup
  }{2015}]{bojanowski2015weakly}
Bojanowski, P.; Lajugie, R.; Grave, E.; Bach, F.; Laptev, I.; Ponce, J.; and
  Schmid, C.
\newblock 2015.
\newblock Weakly-supervised alignment of video with text.
\newblock In {\em IEEE CVPR},  4462--4470.

\bibitem[\protect\citeauthoryear{Buch \bgroup et al\mbox.\egroup
  }{2017}]{buch2017sst}
Buch, S.; Escorcia, V.; Shen, C.; Ghanem, B.; and Niebles, J.~C.
\newblock 2017.
\newblock Sst: Single-stream temporal action proposals.
\newblock In {\em IEEE CVPR},  6373--6382.

\bibitem[\protect\citeauthoryear{Caba~Heilbron \bgroup et al\mbox.\egroup
  }{2015}]{caba2015activitynet}
Caba~Heilbron, F.; Escorcia, V.; Ghanem, B.; and Carlos~Niebles, J.
\newblock 2015.
\newblock Activitynet: A large-scale video benchmark for human activity
  understanding.
\newblock In {\em IEEE CVPR},  961--970.

\bibitem[\protect\citeauthoryear{Chen \bgroup et al\mbox.\egroup
  }{2018}]{chen2018temporally}
Chen, J.; Chen, X.; Ma, L.; Jie, Z.; and Chua, T.-S.
\newblock 2018.
\newblock Temporally grounding natural sentence in video.
\newblock In {\em EMNLP},  162--171.

\bibitem[\protect\citeauthoryear{Devlin \bgroup et al\mbox.\egroup
  }{2019}]{devlin2019bert}
Devlin, J.; Chang, M.-W.; Lee, K.; and Toutanova, K.
\newblock 2019.
\newblock {BERT}: Pre-training of deep bidirectional transformers for language
  understanding.
\newblock In {\em Proceedings of the 2019 Conference of the North {A}merican
  Chapter of the Association for Computational Linguistics: Human Language
  Technologies, Volume 1 (Long and Short Papers)},  4171--4186.
\newblock Minneapolis, Minnesota: Association for Computational Linguistics.

\bibitem[\protect\citeauthoryear{Duan \bgroup et al\mbox.\egroup
  }{2018}]{duan2018weakly}
Duan, X.; Huang, W.; Gan, C.; Wang, J.; Zhu, W.; and Huang, J.
\newblock 2018.
\newblock Weakly supervised dense event captioning in videos.
\newblock In {\em NIPS},  3059--3069.

\bibitem[\protect\citeauthoryear{Escorcia \bgroup et al\mbox.\egroup
  }{2016}]{escorcia2016daps}
Escorcia, V.; Heilbron, F.~C.; Niebles, J.~C.; and Ghanem, B.
\newblock 2016.
\newblock Daps: Deep action proposals for action understanding.
\newblock In {\em ECCV},  768--784.
\newblock Springer.

\bibitem[\protect\citeauthoryear{Gao \bgroup et al\mbox.\egroup
  }{2017}]{gao2017tall}
Gao, J.; Sun, C.; Yang, Z.; and Nevatia, R.
\newblock 2017.
\newblock Tall: Temporal activity localization via language query.
\newblock In {\em IEEE CVPR},  5267--5275.

\bibitem[\protect\citeauthoryear{Hendricks \bgroup et al\mbox.\egroup
  }{2017}]{hendricks2017localizing}
Hendricks, L.~A.; Wang, O.; Shechtman, E.; Sivic, J.; Darrell, T.; and Russell,
  B.
\newblock 2017.
\newblock Localizing moments in video with natural language.
\newblock In {\em IEEE ICCV},  5803--5812.

\bibitem[\protect\citeauthoryear{Hendricks \bgroup et al\mbox.\egroup
  }{2018}]{hendricks2018localizing}
Hendricks, L.~A.; Wang, O.; Shechtman, E.; Sivic, J.; Darrell, T.; and Russell,
  B.
\newblock 2018.
\newblock Localizing moments in video with temporal language.
\newblock In {\em EMNLP},  1380--1390.
\newblock ACL.

\bibitem[\protect\citeauthoryear{Karpathy and Fei-Fei}{2015}]{karpathy2015deep}
Karpathy, A., and Fei-Fei, L.
\newblock 2015.
\newblock Deep visual-semantic alignments for generating image descriptions.
\newblock In {\em IEEE CVPR},  3128--3137.

\bibitem[\protect\citeauthoryear{Liu \bgroup et al\mbox.\egroup
  }{2018}]{liu2018cross}
Liu, M.; Wang, X.; Nie, L.; Tian, Q.; Chen, B.; and Chua, T.-S.
\newblock 2018.
\newblock Cross-modal moment localization in videos.
\newblock In {\em MM},  843--851.
\newblock ACM.

\bibitem[\protect\citeauthoryear{Mithun, Paul, and
  Roy-Chowdhury}{2019}]{mithun2019weakly}
Mithun, N.~C.; Paul, S.; and Roy-Chowdhury, A.~K.
\newblock 2019.
\newblock Weakly supervised video moment retrieval from text queries.
\newblock In {\em IEEE CVPR},  11592--11601.

\bibitem[\protect\citeauthoryear{Nguyen \bgroup et al\mbox.\egroup
  }{2018}]{nguyen2018weakly}
Nguyen, P.; Liu, T.; Prasad, G.; and Han, B.
\newblock 2018.
\newblock Weakly supervised action localization by sparse temporal pooling
  network.
\newblock In {\em IEEE CVPR},  6752--6761.

\bibitem[\protect\citeauthoryear{Otani \bgroup et al\mbox.\egroup
  }{2016}]{otani2016learning}
Otani, M.; Nakashima, Y.; Rahtu, E.; Heikkil{\"a}, J.; and Yokoya, N.
\newblock 2016.
\newblock Learning joint representations of videos and sentences with web image
  search.
\newblock In {\em ECCV},  651--667.
\newblock Springer.

\bibitem[\protect\citeauthoryear{Pennington, Socher, and
  Manning}{2014}]{pennington2014glove}
Pennington, J.; Socher, R.; and Manning, C.
\newblock 2014.
\newblock Glove: Global vectors for word representation.
\newblock In {\em EMNLP},  1532--1543.

\bibitem[\protect\citeauthoryear{Shou \bgroup et al\mbox.\egroup
  }{2017}]{shou2017cdc}
Shou, Z.; Chan, J.; Zareian, A.; Miyazawa, K.; and Chang, S.-F.
\newblock 2017.
\newblock Cdc: Convolutional-de-convolutional networks for precise temporal
  action localization in untrimmed videos.
\newblock In {\em IEEE CVPR},  1417--1426.

\bibitem[\protect\citeauthoryear{Shou, Wang, and
  Chang}{2016}]{shou2016temporal}
Shou, Z.; Wang, D.; and Chang, S.-F.
\newblock 2016.
\newblock Temporal action localization in untrimmed videos via multi-stage
  cnns.
\newblock In {\em IEEE CVPR},  1049--1058.

\bibitem[\protect\citeauthoryear{Sigurdsson \bgroup et al\mbox.\egroup
  }{2016}]{sigurdssonhollywood}
Sigurdsson, G.~A.; Varol, G.; Wang, X.; Farhadi, A.; Laptev, I.; and Gupta, A.
\newblock 2016.
\newblock Hollywood in homes: Crowdsourcing data collection for activity
  understanding.

\bibitem[\protect\citeauthoryear{Singh and Lee}{2017}]{singh2017hide}
Singh, K.~K., and Lee, Y.~J.
\newblock 2017.
\newblock Hide-and-seek: Forcing a network to be meticulous for
  weakly-supervised object and action localization.
\newblock In {\em IEEE ICCV}.

\bibitem[\protect\citeauthoryear{Song \bgroup et al\mbox.\egroup
  }{2019}]{song2019mass}
Song, K.; Tan, X.; Qin, T.; Lu, J.; and Liu, T.-Y.
\newblock 2019.
\newblock Mass: Masked sequence to sequence pre-training for language
  generation.
\newblock {\em arXiv preprint arXiv:1905.02450}.

\bibitem[\protect\citeauthoryear{Tang \bgroup et al\mbox.\egroup
  }{2018}]{tang2018self}
Tang, G.; M{\"u}ller, M.; Rios, A.; and Sennrich, R.
\newblock 2018.
\newblock Why self-attention? a targeted evaluation of neural machine
  translation architectures.
\newblock In {\em EMNLP},  4263--4272.

\bibitem[\protect\citeauthoryear{Vaswani \bgroup et al\mbox.\egroup
  }{2017}]{vaswani2017attention}
Vaswani, A.; Shazeer, N.; Parmar, N.; Uszkoreit, J.; Jones, L.; Gomez, A.~N.;
  Kaiser, {\L}.; and Polosukhin, I.
\newblock 2017.
\newblock Attention is all you need.
\newblock In {\em NIPS},  5998--6008.

\bibitem[\protect\citeauthoryear{Wang \bgroup et al\mbox.\egroup
  }{2017}]{wang2017untrimmednets}
Wang, L.; Xiong, Y.; Lin, D.; and Van~Gool, L.
\newblock 2017.
\newblock Untrimmednets for weakly supervised action recognition and detection.
\newblock In {\em IEEE CVPR}.

\bibitem[\protect\citeauthoryear{Wang, Huang, and
  Wang}{2019}]{wang2019language}
Wang, W.; Huang, Y.; and Wang, L.
\newblock 2019.
\newblock Language-driven temporal activity localization: A semantic matching
  reinforcement learning model.
\newblock In {\em IEEE CVPR},  334--343.

\bibitem[\protect\citeauthoryear{Wang, Li, and Smola}{2019}]{wang190409408}
Wang, C.; Li, M.; and Smola, A.~J.
\newblock 2019.
\newblock Language models with transformers.
\newblock {\em CoRR} abs/1904.09408.

\bibitem[\protect\citeauthoryear{Xia \bgroup et al\mbox.\egroup
  }{2018}]{xia2018model}
Xia, Y.; Tan, X.; Tian, F.; Qin, T.; Yu, N.; and Liu, T.-Y.
\newblock 2018.
\newblock Model-level dual learning.
\newblock In {\em ICML},  5383--5392.

\bibitem[\protect\citeauthoryear{Xu \bgroup et al\mbox.\egroup
  }{2015}]{xu2015jointly}
Xu, R.; Xiong, C.; Chen, W.; and Corso, J.~J.
\newblock 2015.
\newblock Jointly modeling deep video and compositional text to bridge vision
  and language in a unified framework.
\newblock In {\em AAAI}, volume~5, ~6.

\bibitem[\protect\citeauthoryear{Xu \bgroup et al\mbox.\egroup
  }{2019}]{xu2019multilevel}
Xu, H.; He, K.; Sigal, L.; Sclaroff, S.; and Saenko, K.
\newblock 2019.
\newblock Multilevel language and vision integration for text-to-clip
  retrieval.
\newblock In {\em AAAI}, volume~2, ~7.

\bibitem[\protect\citeauthoryear{Zhang \bgroup et al\mbox.\egroup
  }{2019a}]{zhang2019man}
Zhang, D.; Dai, X.; Wang, X.; Wang, Y.-F.; and Davis, L.~S.
\newblock 2019a.
\newblock Man: Moment alignment network for natural language moment retrieval
  via iterative graph adjustment.
\newblock In {\em IEEE CVPR},  1247--1257.

\bibitem[\protect\citeauthoryear{Zhang \bgroup et al\mbox.\egroup
  }{2019b}]{zhang2019cross}
Zhang, Z.; Lin, Z.; Zhao, Z.; and Xiao, Z.
\newblock 2019b.
\newblock Cross-modal interaction networks for query-based moment retrieval in
  videos.
\newblock In {\em ACM SIGIR},  655--664.

\bibitem[\protect\citeauthoryear{Zhao \bgroup et al\mbox.\egroup
  }{2017}]{zhao2017temporal}
Zhao, Y.; Xiong, Y.; Wang, L.; Wu, Z.; Tang, X.; and Lin, D.
\newblock 2017.
\newblock Temporal action detection with structured segment networks.
\newblock In {\em IEEE ICCV}.

\end{thebibliography}
\bibliographystyle{aaai}

\end{document}